\documentclass[letterpaper]{article} 
\usepackage[]{aaai23}
\usepackage{times}  
\usepackage{helvet}  
\usepackage{courier}  
\usepackage[hyphens]{url}  
\usepackage{graphicx} 
\usepackage{natbib}  
\usepackage{caption} 
\usepackage{algorithm}
\usepackage{algorithmic}
\usepackage{amssymb}
\usepackage{amsmath}
\usepackage{colortbl}
\usepackage[table]{xcolor}
\usepackage{array}
\usepackage{diagbox}
\usepackage{makecell}
\usepackage{multirow}
\usepackage{newfloat}
\usepackage{listings}
\DeclareCaptionStyle{ruled}{labelfont=normalfont,labelsep=colon,strut=off} 
\floatstyle{ruled}
\newfloat{listing}{tb}{lst}{}
\floatname{listing}{Listing}
\nocopyright

\title{\textsc{PromptFL}: Let Federated Participants Cooperatively Learn Prompts Instead\\ of Models — Federated Learning in Age of Foundation Model}
\author{
    Tao Guo, Song Guo, Junxiao Wang, Wenchao Xu
}
\affiliations{

    Department of Computing, The Hong Kong Polytechnic University, Hong Kong, China
}

\usepackage{bibentry}

\begin{document}
\maketitle

\begin{abstract}
Quick global aggregation of effective distributed parameters is crucial to federated learning (FL), which requires adequate bandwidth for parameters communication and sufficient user data for local training. Otherwise, FL may cost excessive training time for convergence and produce inaccurate models. In this paper, we propose a brand-new FL framework, \textsc{PromptFL}, that replaces the federated model training with the federated prompt training, i.e., let federated participants train prompts instead of a shared model, to simultaneously achieve the efficient global aggregation and local training on insufficient data by exploiting the power of foundation models (FM) in a distributed way. \textsc{PromptFL} ships an off-the-shelf FM, i.e., CLIP, to distributed clients who would cooperatively train shared soft prompts based on very few local data. Since \textsc{PromptFL} only needs to update the prompts instead of the whole model, both the local training and the global aggregation can be significantly accelerated. And FM trained over large scale data can provide strong adaptation capability to distributed users tasks with the trained soft prompts. We empirically analyze the \textsc{PromptFL} via extensive experiments, and show its superiority in terms of system feasibility, user privacy, and performance.


\end{abstract}

\section{Introduction}
The ever-growing edge devices, e.g., smart phones, autonomous vehicles, etc., are generating various types and rapidly growing big data. Artificial intelligence (AI) has shown its success to mine the big edge data and produce accurate models that can replace human decisions timely and properly. Traditional AI paradigms require to gather all raw data to a cloud center for centralized training, which can incur significant communication overhead and potential privacy leakage, and thus are not desirable for edge users. 
    

%

Federated learning (FL) has emerged to conduct distributed machine learning that allows multiple edge users to jointly train a shared model without sharing their raw data, which has been demonstrated great success in many edge applications, e.g., input word prediction, voice assistant, etc. \cite{hard2018federated,liang2020think}, that can mine massive distributed data without exposing users' privacy, and thus are widely applied in various edge scenarios. The FL training process comprises of two iterative phases, i.e., local training and global aggregation. Thus the learning performance is determined by both the effectiveness of the parameters from local training and smooth aggregation of them. However, these two requirements are not easy to satisfy in edge environment, i.e., edge users often have limited bandwidth and insufficient data, which can cause inefficient parameters aggregation, excessive training time and reduced model accuracy.  


Existing research efforts have focused on improving the FL optimization process \cite{li2020federated,zhao2018federated} or refining model architectures \cite{qu2022rethinking}, but this does not change that FL inherently entails a large number of communication rounds and a large amount of labeled data for training, which are often unavailable for edge users.
Such challenges are particularly salient under the combined effect of a long training process and unfavorable factors such as non-IID and unbalanced data, limited communication bandwidth, and unreliable and limited device availability. 
%

%

We revisits the question of how FL mines the distributed data in iterative training rounds, and exploit the emerging foundation model (FM) to optimize the FL training. FM refers to large neural model that trained on large scale data and has strong adaptation capability for various downstream tasks. We let federated participants cooperatively learn prompts instead of models to unleash the power of FM in a distributed way, whereby both the local training and global aggregation can be significantly accelerated.    
   
%
%
%

We investigate the behavior of the nascent model in a standard FL setting using popular off-the-shelf FMs, \emph{e.g.}, CLIP, and methods for FM adaptation. 
We propose \textsc{PromptFL}, a framework that replaces existing federated model training with prompt training, i.e., FL clients train prompts instead of a model, which can simultaneously exploit the insufficient local data and reduce the aggregation overhead.
%
\textsc{PromptFL} ships an off-the-shelf public CLIP to users and apply continuous prompts (\emph{a.k.a.} soft prompts) for FM adaptation, which requires very few data samples from edge users.
The framework is technically very simple but effective.
The focus of our investigation is whether it meets the key principles:
\begin{itemize}
  \item \textbf{Feasibility.} What are the system costs? We examine the feasibility of \textsc{PromptFL} on modern hardware, focusing conservatively on personal cell phones. We demonstrate the feasibility of the system in terms of overhead in communication, training, and inference dimensions.
  
  \item \textbf{Performance.} Are \textsc{PromptFL} competitive with FL? FL does not baseline against any such approach, so we implement a proof-of-concept in the framework, spanning a range of popular image classification tasks. We observe \textsc{PromptFL} competitive with strong FL baselines.
  
  \item \textbf{Privacy.} Is \textsc{PromptFL} privacy-preserving? We show that \textsc{PromptFL} keeps data on each device private, aiming to learn global prompts updated only by communicating gradients rather than the data itself, and thus not less private than FL.
\end{itemize}

\section{Preliminaries}
\subsection{Foundation Model}
AI is going through a paradigm shift with the rise of models (\emph{e.g.}, BERT, GPT-3, CLIP, DALL-E$\cdot$2) trained on broad data using self-supervision at scale that can be adapted to a wide range of downstream tasks.
Researchers call these models foundation models (FMs) to emphasize their key core.
From a technical standpoint, FMs are not new.
However, the sheer size and scope of FMs over the past few years has expanded our imagination of what is possible.
%
%
%
%
FMs are scientifically interesting for their impressive performance and capabilities, but what makes them critical to research is that they are rapidly being integrated into real-world deployments of AI systems, with profound implications for users.

\paragraph{CLIP}
Contrastive Language-Image Pre-Training (CLIP) is a neural network trained on hundreds of millions of (image, caption) pairs \cite{radford2021learning}.
CLIP encodes images and captions separately as vectors, enabling users with visual modality samples to retrieve, score, or classify samples from textual modalities.
Models are often very fragile and only know very specific things you trained them to do. CLIP extends the knowledge of classification models to a wider range of things by leveraging semantic information in text. 
Standard classification models completely discard the semantic meaning of class labels and simply enumerate numeric classes behind the scenes; CLIP works by understanding the meaning of the classes.
ALIGN is another CLIP-like vision-language pre-training \cite{jia2021scaling}.

\paragraph{Image Classification with CLIP} CLIP pre-trains an image encoder and a text encoder to predict which images are paired with which texts.
We can use this behavior to convert the CLIP to an image classifier.
We may convert all [class] to captions such as ``picture of [class]'' and predict the caption class CLIP estimates the best pairing with the given image.
In many previous works, this has involved prompt template engineering, in which human engineers or algorithms search for the best template for each class \cite{furst2021cloob,li2021supervision,singh2022flava,yuan2021florence}.

\subsection{Federated Learning}
Recent neural models require large amounts of training data \cite{dodge2020fine}, and users typically hold limited-scale labeled data.
To address the challenge of lack of sufficient data for individual users, federated learning of data across multiple privacy spheres (\emph{i.e.}, users) has become a popular framework.

The term \emph{federated learning} was introduced by \cite{mcmahan2017communication}.
In a centralized setting, the federated server initially sends global model parameters to each client. 
After training with local data, the participants are only required to share gradients for model updates. 
Then the server aggregates the gradients and transmits the updated model back to each client. 
More specifically, federated learning is a machine learning setting where a set of $n$ clients (\emph{e.g.,} mobile devices) collaboratively train a model under the orchestration of a federated server (\emph{e.g.}, service provider), while the training data of clients is stored locally and not exchanged \cite{kairouz2021advances}. 
The federated server orchestrates the collaborative training process, by repeating the following steps until training is converged:

\paragraph{Client Selection}
Given the unstable client availability, for the round $t$ of federated learning, the federated server samples a small subset of $m$ clients meeting eligibility requirements out of all $n$ clients to participate in the learning.

\paragraph{Local Training} 
Upon notification of being selected at the round $t$, each selected client downloads the current parameters $\theta$ of global model and a training program from the federated server.
Each selected client locally computes an update to the global model on its local training data by executing the training program. 
More specifically, the gradients updated at one client (denoted as $G$), are computed by $\frac{\partial\ell(X,y,\theta)}{\partial\theta}$, where $X$, $y$ denote the batches of training data and corresponding labels, and $\ell(\cdot)$ refers to the loss function. 

The gradients $G$ in typical federated learning settings are the minimum that must be shared to the server, corresponding to FedSGD method. 
In FedAVG \cite{mcmahan2017communication}, models are consecutively updated on more batches of local data, which can be several epochs of training, and then shared.
We note that a common way is to share the updated model $\theta+G$, but this practically amounts to sharing $G$ since all participants know $\theta$.

\paragraph{Global Aggregation}
Upon having received local updates from $m$ clients, the federated server aggregates these updates and update its global model, and initiates next round learning.
In addition to the federated learning framework that relies on the centralized server node, there are also some federated learning implementations based on the decentralized framework \cite{roy2019braintorrent,lalitha2018fully,hu2019decentralized}.
This means that the aggregation of gradients does not necessarily occur in a fixed federation server, but may also occur in some clients.

\begin{figure*}[t]
\setlength{\abovedisplayskip}{0pt}
\setlength{\belowdisplayskip}{0pt}
\centering
\includegraphics[width=0.75\linewidth]{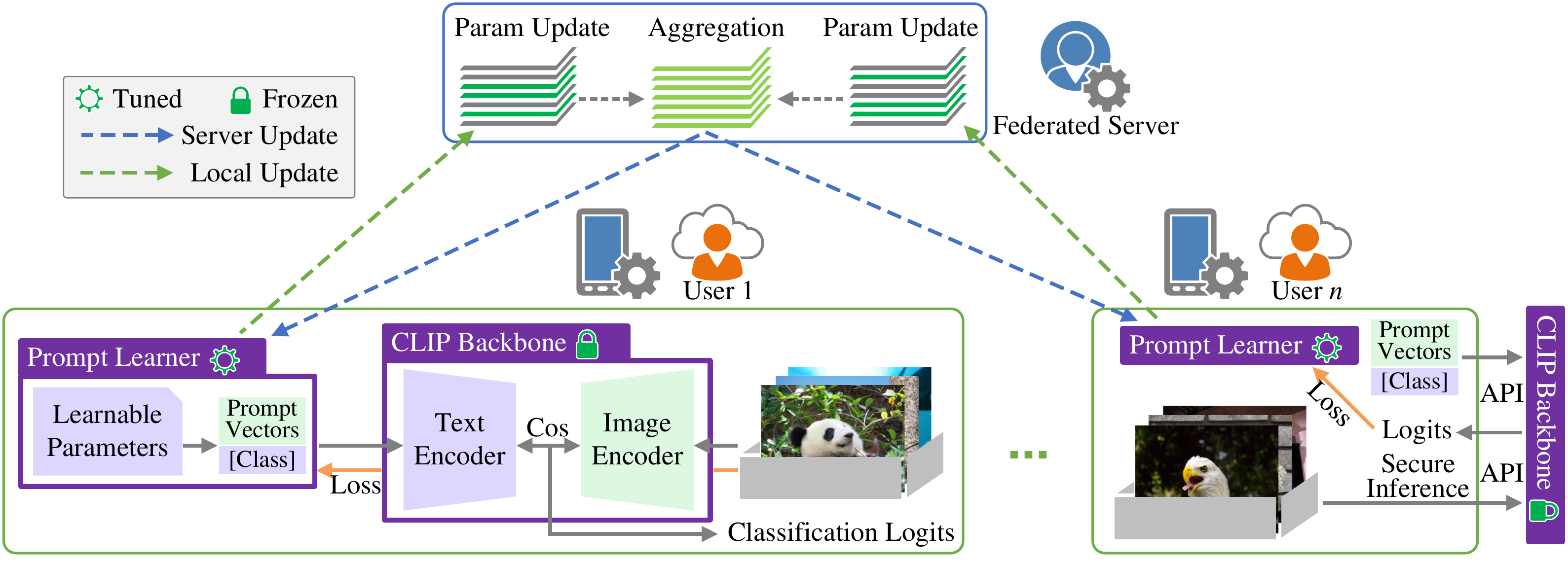}
\caption{Framework and workflow of \textsc{PromptFL}. Each client includes a prompt learner (with only a small amount of trainable parameters) and an out-of-the-box CLIP (with backbone frozen). The federated server aggregates only the parameter updates of prompt learners over multiple users, and transmit the updated parameters back to each user.}
\label{fig:framework}
\end{figure*}

\section{Prompt-Based Federated Learning}
We hypothesize that an off-the-shelf public CLIP-like model is shipped to the user device. 
The CLIP-like model is a powerful image classifier that utilizes linguistic knowledge to classify images.
In other words, CLIP already knows a lot about the content of images.
But to unleash the power of CLIP in FL, we need to take advantage of something called prompt engineering that was mentioned in the preliminaries.

\subsection{Prompt Engineering}
The prompting function $f_{\text{prompt}}(\cdot)$ is applied to modify the class label $\textbf{y}$ into a prompt $\textbf{y}'=f_{\text{prompt}}(\textbf{y})$.
The most natural form of implementing a prompting function is to manually create an intuitive template based on human introspection.
For example, as referred in \cite{brown2020language} we may manually craft prefix prompts to handle an image classification task by using templates like ``picture of [class]'' or ``a photo of a [class]''.
Based on that, many approaches have been proposed to automate the template design process.

Specifically, the automated prompting can be further separated into discrete prompts (\emph{a.k.a.} hard prompts), where the prompt is an actual text string, and continuous prompts (\emph{a.k.a.} soft prompts), where the prompt is performed directly in the embedding space of the model \cite{liu2021pre}.
Discrete prompts constraint that the embeddings of template words be the embeddings of natural language words \cite{shin2020autoprompt,gao2021making}.
Thus, discrete prompting is a clear way to visualize what ``word'' are learned for the vectors \cite{deng2022rlprompt}.

Our paper adopts continuous prompts instead of discrete prompts in FL for the reason that (1) Our purpose of prompt construction is to find a way to enable FL to efficiently perform the image classification tasks, not for human interpretation, there is no need to limit prompts to human-interpretable natural language. 
(2) The templates have their own parameters that can be tuned based on training data from the user, which is a natural compatibility connecting FL and prompting.
More related topics of continuous prompts can refer to \cite{li2021prefix,lester2021power,tsimpoukelli2021multimodal,hambardzumyan2021warp,zhou2021learning}.

\subsection{Framework to Learn Prompts in FL}
The framework of \textsc{PromptFL} is presented in Figure~\ref{fig:framework}.
Each FL client consists of a prompt learner and an out-of-the-box CLIP model.
\textsc{PromptFL} introduces only a small amount of trainable parameters in the prompt learner while keeping the CLIP backbone frozen.
In other words, during local training, only the parameters of the prompt learner are updated while the whole CLIP model turns off gradients in both the image and the text encoder.
The federated server is designed to aggregate only the parameter updates of prompt learners over multiple users, and transmit the updated parameters back to
each user. 
Thus, \textsc{PromptFL} evolves the goal of FL from model training to prompt learner training.

The CLIP backbone comprises two encoders, one for images and the other for texts. 
The image encoder will map high-dimensional images into a low-dimensional embedding space.
The network of the image encoder can take the form of a CNN such as ResNet50 \cite{he2016deep} or Vision Transformer \cite{dosovitskiy2021image}.
The text encoder will generate text
representations from input.
The network of the text encoder is a Transformer \cite{vaswani2017attention}.

\begin{table*}[t]
  \centering
  \rowcolors{2}{blue!20}{blue!10}     
  \begin{tabular}{|c|c|c|c|}
  \hline
  \rowcolor{red!10}
  \hline
    \diagbox{Dimensions}{Frameworks} & \makecell[c]{\textsc{PromptFL} \\ (150M parameter model)} & \makecell[c]{Federated Learning \\ (100M parameter model)} & \makecell[c]{Modern Mobile Phone Hardware \\ \cite{andrew2021apple}} \\
    \hline 
    Communication & \makecell[c]{600 MB File Download \ \ \\ 1.4 Minutes} & \makecell[c]{40 GB File Download + \ \ \\ 40 GB File Upload \\ Totally 9 Hours} & \makecell[c]{54 Mbps Downstream RateLimit \\ 12 Mbps Upstream RateLimit \\ \cite{dea2021average}} \\
    \hline 
    Training & Almost None & 4 TFLOPs & 1.5 TFLOPs, 8 GB RAM \\
    \hline 
    Inference & 60 GFLOPs & 40 GFLOPs & 1.5 TFLOPs, 8 GB RAM \\
    \hline
    Storage & 600 MB on Disk & 400 MB on Disk & 1 TB on Disk \\
    \hline
  \end{tabular}
  \caption{System cost comparison. Assumes 32 local training batch size, 1 local training epoch, 100 total communication rounds for FL. Assumes 196 input sequence length, full precision for \textsc{PromptFL} and FL.}
  \label{tab:cost}
\end{table*}

\paragraph{Prompt Learner}
Given a pre-trained CLIP backbone, the input to the text encoder is designed in the form of [prompt vectors][class].
Inspired by the simple and straightforward prompt design in \cite{zhou2021learning},
we introduce a set of $p$ continuous embeddings of dimension $d$ in the [prompt vectors].
$d$ is same as the dimension of word embeddings in the text encoder, thus 512 by default.
$p$ is a hyperparameter specifying the number of embeddings.
In a word, [prompt vectors] are $p$ learnable $d$-dimensional vectors.

Given a batch of image-text pairs, CLIP will maximize the cosine similarity for matched pairs while minimize the cosine similarity for all other unmatched pairs.
Since CLIP is pre-trained to predict whether an image matches a textual description, it can compute the classification loss and logits by aligning the two embedding spaces for images and texts (\emph{i.e.}, [prompt vectors][class]) respectively.
Formally, let $g(\cdot)$ and $h(\cdot)$ be the feature extraction function of the image and text encoder.
Let $w_i=h(\textbf{P}, \textbf{K}_i)$ be the weight vector generated by the text encoder, where $i\in[1,k]$.
$k$ denotes the number of classes and each $(\textbf{P}, \textbf{K}_i)$ is derived from the prompt in the form of [prompt vectors][class]$_i$, where [class]$_i$ is replaced by the word embedding vector of specific class label name.
Let $\cos[\cdot|\cdot]$ denote the cosine similarity used by CLIP.
By forwarding a $(\textbf{P}, \textbf{K}_i)$ and an image $\textbf{x}$,  the classification prediction probability and logits are computed as 
\begin{align} 
    \text{p}(\textbf{y}=i|\textbf{x})=\frac{\exp\left(cos[g(\textbf{x})|h(\textbf{P},\textbf{K}_i)]\right)}{\sum_{j=1}^{k}\exp\left(cos[g(\textbf{x})|h(\textbf{P},\textbf{K}_j)]\right)},
\end{align}
where $\textbf{P}$ is the only part that is updated in local back propagation and aggregated in the federated server.

Prompting are particularly useful in the FL case, as using prompts to push the model in the correct direction is particularly effective.
This feature enables prompting to converge quickly in FL, requires less data per user, and is less affected by adverse factors in the process, \emph{e.g.}, non-IID and unbalanced data, limited communication bandwidth, and unreliable and limited device availability.
In this paper, the prompt learner employed in \textsc{PromptFL} though simple and straightforward as a bridge to our core idea is easy to follow.
We also envision that more complex and effective bridges would be there to replace the role and should be a valuable direction.

\subsection{System Feasibility}
We examine the feasibility of \textsc{PromptFL} on modern hardware, focusing conservatively on personal cell phones.
We notice that users can access GPUs from their mobile phones.
Enterprise users have more abundant resources.
Without loss of generality, we take a 100M parameter model for FL and 150M parameter CLIP backbone for image similarity-search of \textsc{PromptFL}.
The prompt learner introduces only a small number of parameters, that can be ignored.
We assume that the FL configures 32 local training batch size, 1 local training epoch, and 100 total communication rounds, which suggested in \cite{qu2022rethinking}.
We also assume that both FL and \textsc{PromptFL} configure 196 input sequence length and the full precision.
The system cost comparison is summarized in Table~\ref{tab:cost} along the following dimensions:

\paragraph{Communication}
The average download speed within the globe for mobile internet was 54 Mbps, and the average upload speed for mobile internet was 12 Mbps that reported by 2021 \cite{dea2021average}.
\textsc{PromptFL} requires locally downloading while FL requires communicating the model repeatedly between users and the federated server.
Thus, the communication cost in terms of file transfer volume is that it takes only 1.4 minutes to transfer 600MB for \textsc{PromptFL}, and 9 hours for FL to transfer 40GB.

\begin{table*}[!ht]
\renewcommand\arraystretch{1.1}
\centering
\begin{tabular}{l l l l l l l l l l l l }
\Xhline{3\arrayrulewidth}
\multirow{3}*{\textsc{Benchmark}}&\multirow{3}*{\textsc{Method}}&\multicolumn{4}{c}{\textsc{IID}} &\multicolumn{4}{c}{\textsc{Extreme Non-IID}}& \multicolumn{2}{c}{\textsc{Learnable}} \\
~& & \multicolumn{2}{c}{\textit{Accuracy}$\uparrow$}& \multicolumn{2}{c}{\textit{F-1}$\uparrow$}& \multicolumn{2}{c}{\textit{Accuracy}$\uparrow$}&\multicolumn{2}{c}{\textit{F-1}$\uparrow$}& \multicolumn{2}{c}{\textsc{Parameters}}  \\  

\cline{3-12}

~& & Rn50& Vit &Rn50 & Vit& Rn50&Vit&Rn50&Vit&Rn50&Vit\\\Xhline{3\arrayrulewidth}

\multirow{3}*{Caltech101}&\textsc{PromptFL} &\textbf{90.18} &\textbf{94.65}&\textbf{86.09} &\textbf{91.76} &\textbf{88.72}&\textbf{94.12} &\textbf{83.98}  & \textbf{90.48}&0.1\% & 0.01\%   \\  
~&Finetuning FL&90.02 &93.1&84.72 & 89.07 &29.78& 29.89 &12.2  & 12.2 & 100\% & 100\%\\ 
~&FL from scratch&32.41&32.49  &10.51 &12.89 &- & - &- &- & 100\% & 100\%\\\hline
   
\multirow{3}*{Flowers102}&\textsc{PromptFL}&88.14& 90.5 &87.62 &90.14 &\textbf{66.3} &\textbf{74.75}&\textbf{60.14} &\textbf{69.13} & 0.1\% & 0.01\%\\  
~&Finetuning FL&\textbf{92.6} &\textbf{91.9} &\textbf{91.56} &\textbf{90.7} &24.4&24.5 &10.68 &11.18  & 100\% &100\% \\ 
~&FL from scratch&33.17&38  &25.7 &32.5 &- & - &- &- &100\% & 100\%\\\hline

\multirow{3}*{OxfordPets}&\textsc{PromptFL} &88.5&\textbf{92.89} &88.44  & \textbf{92.8} &\textbf{87.03}&\textbf{89.51} &\textbf{86.85} &\textbf{88.45} & 0.1\% & 0.01\%  \\  
~&Finetuning FL &\textbf{90.38}&92.1 &\textbf{90.06}  &91.92 &24.83&25.27 & 11.3  &11.93 & 100\%&100\% \\ 
~&FL from scratch &10.25&8.722 &7.624 &8.318 &- &- &- &- & 100\% & 100\%\\\hline
   
\multirow{3}*{Food101}&\textsc{PromptFL} &\textbf{78.0}&\textbf{85.75} & \textbf{77.9} &\textbf{85.66} &\textbf{78.1}& \textbf{85.88} &\textbf{78.03} &\textbf{85.8} & 0.1\% & 0.01\% \\  
~&Finetuning FL& 69.28&76.68&69.08 &76.85 &22.92&23.8 & 10.19 &10.73 & 100\% & 100\%\\ 
~&FL from scratch&21.11&21.03  &19.75 &19.92 &- &-  &- &- & 100\% &100\% \\\Xhline{3\arrayrulewidth}
\end{tabular}
\caption{\textbf{Performance of \textsc{PromptFL} against existing FL framework on the four datasets}. The table report the accuracy and F-1 score according to the corresponding backbone and method. The best score of each group appears in bold. Compared with finetuning and training from the scratch, \textsc{PromptFL} only update 0.01\% $\sim$ 0.1\% parameters, however, still outperforms other methods across datasets. Given the poor result in training from the scratch even with iid mode, we assume that the performance from the non-iid setting can be even wore, so we omit the result in this row.}
\label{tab:mainresult}
\end{table*}

\paragraph{Training and Inference}
FL requires FLOPs computed by  (2$\times$3$\times$model parameters$\times$local training epoch$\times$local training batch size$\times$input sequence length) for training, while the training FLOPs of \textsc{PromptFL} is much smaller and negligible compared to FL.
For both \textsc{PromptFL} and FL, inference requires FLOPs computed by (2$\times$model parameters$\times$input sequence length), in the setting where the key and value vectors for attention computation are cached.
Compared to the acceptable computational and storage costs, the RAM on the modern cell phones is a key bottleneck.
We believe that this bottleneck will no longer be a problem in the near future as the techniques evolve: (1) Out-of-the-box offloading inference \cite{rajbhandari2021zero}. 
(2) Trends for more RAM \cite{patterson2022black} and tiny CLIPs \cite{sisodia2021distillation}.
(3) Inference with quantization methods \cite{gholami2021survey}.

\paragraph{Compatibility}
Apart from image classification, many different vision tasks are compatible with \textsc{PromptFL}, such as object detection \cite{gu2021open}, video understanding \cite{xu2021videoclip} and visual question answering \cite{shen2021much}.
This means that the system cost of \textsc{PromptFL} is shared by many tasks.
The prompt learner incurs these costs per personal task specific user subset requires.
\textsc{PromptFL} is thus competitive in terms of economics.

\subsection{Privacy Concerns}
As we have outlined in the framework, \textsc{PromptFL} achieves to train prompts in concert with the federated server. 
Each participant user only needs to upload its local parameter update of the prompt learner rather than the raw data of images.
Such a method avoids leakage of raw images, thereby better adapting to the privacy-preserving settings of the FL.
On the other hand, the parameters of prompt learner only describes the correlation between classes and textual prompts, and do not directly contain any visual feature embeddings.
Also, the parameters of prompt learner are static (\emph{i.e.}, input-agnostic) across the training data.
This is useful when faced with a server that wants to recover the raw data from an update \cite{zhu2019deep}.

\paragraph{Inference APIs} 
While pre-trained CLIPs are available for download at the time of writing this paper, high-performance models in these domains are often costly to train.
For example, the CLIP model trained on 400 million labeled images. 
The training process took 30 days across 592 V100 GPUs \cite{radford2021learning}. 
This would have cost million dollars to train on AWS on-demand instances.
The value of these models and their exposure over publicly-accessible APIs make us rethink the framework of \textsc{PromptFL}.
As illustrated in Figure~\ref{fig:framework}, we hypothesize that the model APIs typically return low-dimensional outputs like confidence scores or logits, so information leakage is significantly reduced \cite{dziedzic2022difficulty}.
In such a case, the prompt learner can still be trained normally, because the CLIP backbone is kept frozen during the training process.
The difference is that users need to make queries to the model APIs with their private images.
Some lightweight secure inference techniques like \cite{liu2020datamix} can be used in the framework to protect privacy.

\begin{figure*}[t]
  \centering
  \includegraphics[width=0.85\linewidth]{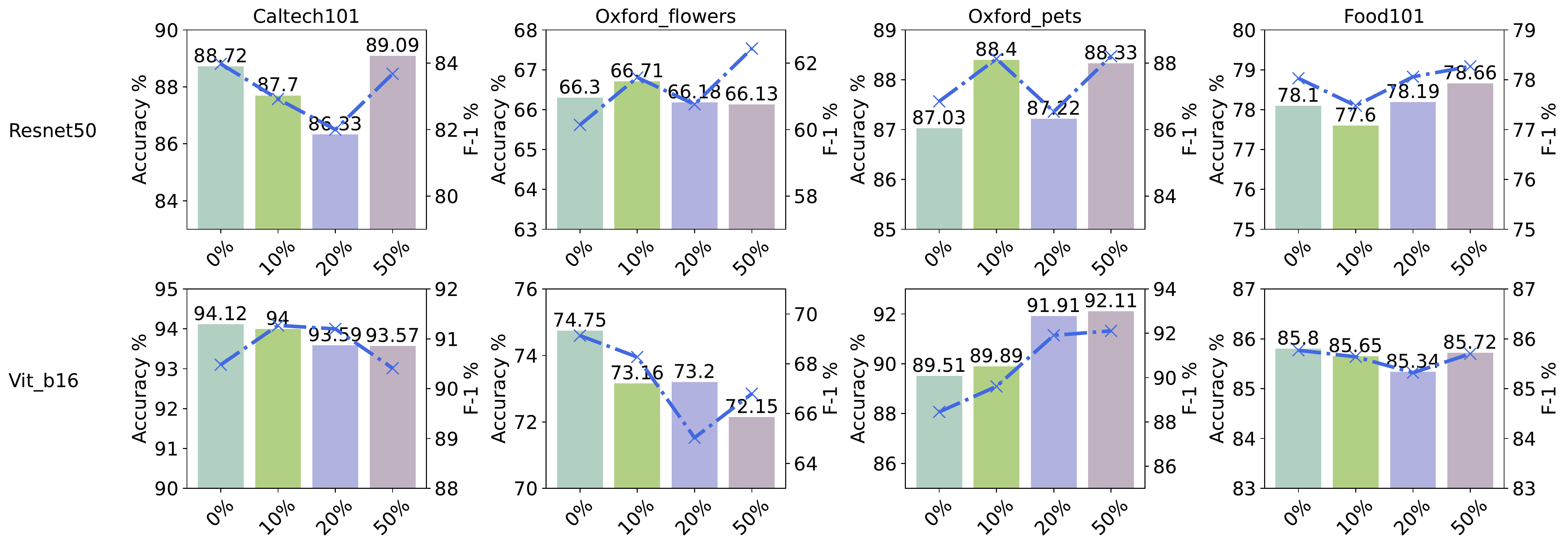}
  \caption{\textbf{Performance of \textsc{PromptFL} with different class distribution.} Bars represent accuracy and lines indicate F-1 score. We range the class distributed on each client from entirely disparate to 10\%, 20\% and 50\% number of classes repeated on more than one client. Compared with the collapse of existing framework in ~\ref{tab:mainresult}, the performance of \textsc{PromptFL} remains stable and competitive. Further more, 50\% overlapping of classes shows slightly improvement across majority of datasets and backbone.}
  \label{fig:repeat}
\end{figure*}

 \begin{figure}[t]
  \centering
  \includegraphics[width=0.85\linewidth]{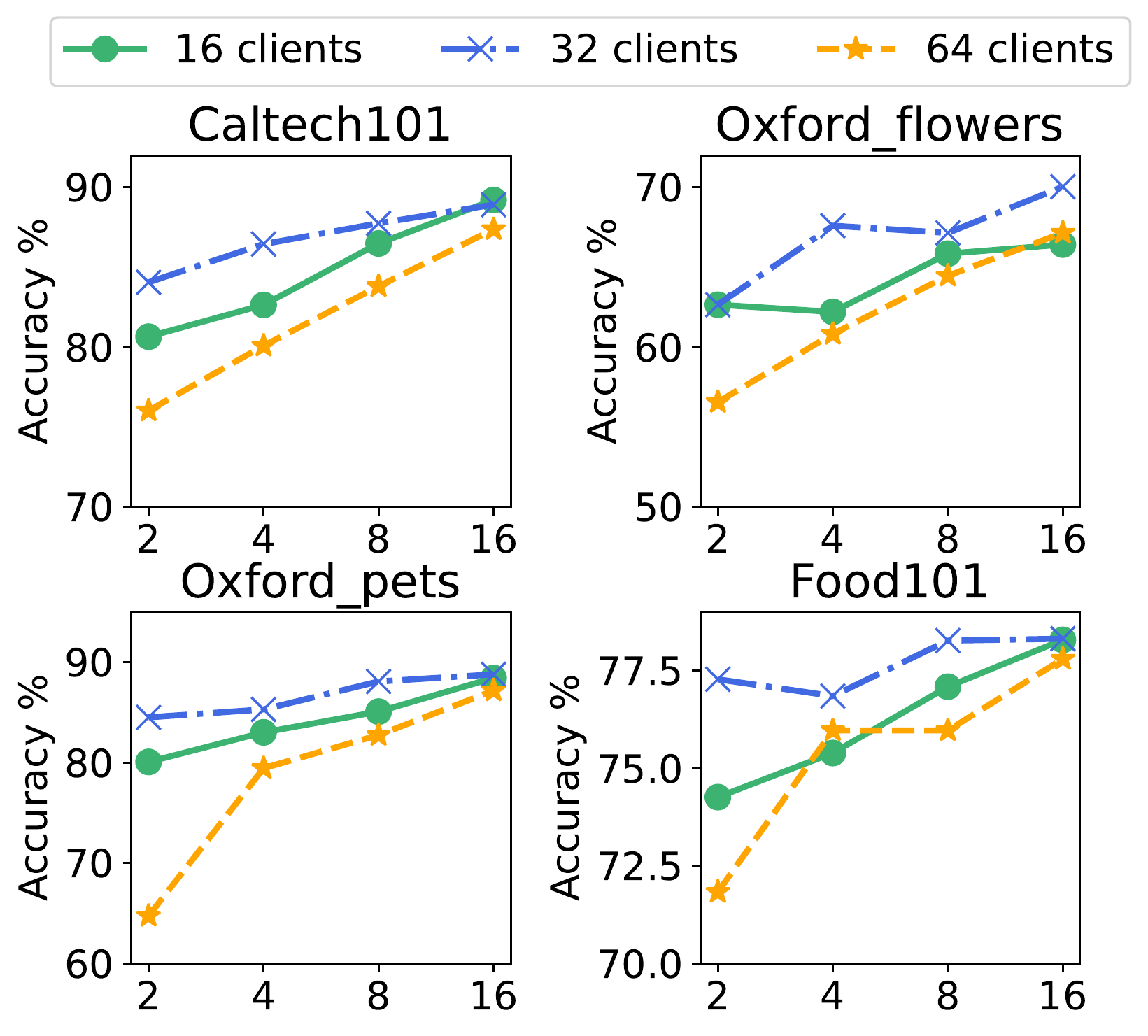}
  \caption{\textbf{Performance of \textsc{PromptFL} with different clients and shots.} The overall performance enhanced as the number of shots increasing. However, as the classes on each client become sufficient, the performance of clients with different clients reach similar optimal results, which on the other hand reveals that clients number do not affect the performance of \textsc{PromptFL}.}
  \label{fig:line}
\end{figure}

\begin{figure*}[t]
  \centering
  \includegraphics[width=0.8\linewidth]{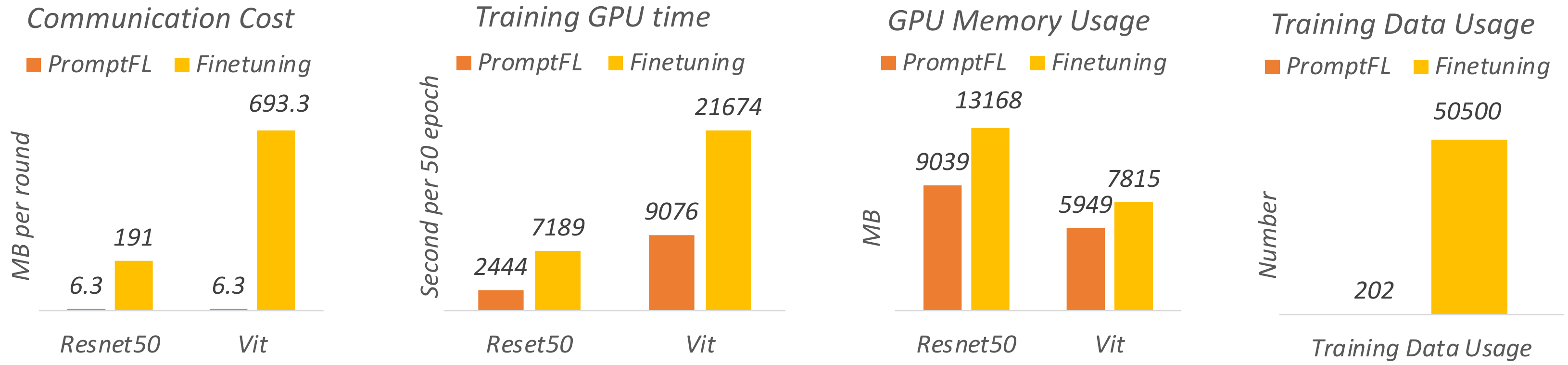}
  \caption{\textbf{Comparison of computation and communication cost of \textsc{PromptFL} and Finetuning FL.} We measure the communication cost by the size of uploaded data per round, and observe that finetuning FL takes up to 110 times of cost more than \textsc{PromptFL}. Furthermore, finetuning and training from scratch take 2 to 3 times of round more than \textsc{PromptFL} for training, which exacerbate the communication expenses. We also utilize GPU memory usage, training GPU time and training data usage to evaluate the computational cost. Training GPU time is calculated by the time of training 50 epoch and training data usage is reported by training food101, which we can observe that finetuning require 250$\times$ more than \textsc{PromptFL}. We can see that \textsc{PromptFL} surpasses the existing framework in the entire aspects of communication and computation efficiency.}
  \label{fig:cost}
\end{figure*}

\section{Experiments}
Our experiments aim to answer the following research questions that are important for the practical deployment of FL methods, while also contributing to our understanding of the \textsc{PromptFL} paradigm.
\begin{itemize}
  \item Is \textsc{PromptFL} able to train a competitive performance in FL as compared to  which have been the de-facto method on image classification tasks?
  
  \item Is \textsc{PromptFL} capable of handling heterogeneous data distributions (\emph{a.k.a.} non-IID settings) across clients?
  
  \item Is \textsc{PromptFL} competitive with the de-facto method in terms of computational communication overhead?
  
  \item What is the difference between \textsc{PromptFL} and the fine-tuning of visual pre-trained models in FL?
  
  \item What practical tips help the service provider and participants deploy \textsc{PromptFL} in FL?
\end{itemize}

%

\subsection{Experimental Setup}
\paragraph{Datasets} 
We select a representative collection of recognition datasets used in CLIP as our benchmarks. 
\textbf{General Objects:} Caltech101 \cite{fei2004learning} for general object detection. \textbf{Fine-grained Categories:} Flowers102 \cite{nilsback2008automated}, OxfordPets \cite{parkhi2012cats} and Food101 \cite{bossard2014food} for fine-grained classification from diversified categories.

\paragraph{Baselines} 
As compared to our proposed \textsc{PromptFL}, we choose current representative framework in FL, FedAVG, by updating and averaging the model weights collaboratively among server and clients. 
We compare both training from the scratch and fine-tuning with pretrained models as our baseline method. 
We select the most prevailing models, Vit\underline{\space}b16 and Retnet50, as our backbone in both our image encoder of \textsc{PromptFL} and the corresponding backbone in the baseline method.

\paragraph{Fine-tuning \emph{vs.} Prompting}
How does the prompting differ from the existing adaptation method in FL?
Currently in vision, the standard adaptation method is fine-tuning.
Therefore we consider fine-tuning as the de-facto way of adapting visual pre-trained models in FL.
Fine-tuning is highly flexible in its usage: it can adapt the pre-trained models to new input domains or new tasks with different output semantics.
Yet it also requires some level of access to the pre-trained models: often entire parameters.
Unlike fine-tuning, prompting adapts the inputs to a pre-trained model by modifying the model's inputs. 
This opens up unique applications: the input-space adaptation puts control in the hands of the FL user; FL users only need to find the prompts, they don't need to control the pre-trained model itself while training and testing.
In this way, FL users can provide adapted images and prompts to an online API that can only operate on their inputs.
On the other hand, fine-tuning is typically conditioned on inputs. Its update also directly contains some embeddings of visual feature information. 
In contrast, the prompts we explore in this paper are input-agnostic across the training data.
So the prompting can prevent leaking of user's private information from FL update to a certain extent.

\paragraph{CLIP \textsc{PromptFL}}
For CLIP, an image-language model, \textsc{PromptFL} organizes users to collaboratively learn prompts as the CLIP's output transformation function.
Given a frozen pre-trained CLIP $\mathcal{F}$ and a task dataset $\mathbf{D}\{(\textbf{x}_m,\textbf{y}_m)\}$ across clients, the target of \textsc{PromptFL} is to learn a single, static, task-specific prompting $f_{\text{prompt}}$ on class space parameterized by [prompt vectors].
Image classes are represented by labels (\emph{e.g.}, `panda') which are then prompted (\emph{i.e.}, `[prompt vectors][panda]') to specify the context of the user's task.
We follow CLIP's protocol and compute the cosine similarity of the embeddings for each class, normalized to a probability distribution via softmax. 
The class with the highest probability is selected as the model output.
The prompting is added to the class space to form a prompted output $\textbf{y}+v_f$.
During training, \textsc{PromptFL} will maximize the likelihood of the correct label $\textbf{y}$,
\begin{align} 
    \max_{f_{\text{prompt}}}\text{p}_{\mathcal{F};f_{\text{prompt}}}(\textbf{y}+v_f|\textbf{x}),
\end{align}
while the gradient updates are applied only to the [prompt vectors] $v_f$ and the CLIP parameters $\mathcal{F}$ remain frozen.
During validation, the optimized prompt is added to all test-time classes, $\mathbf{D}_{\text{test}}\{(\textbf{x}_m,\textbf{y}_m+v_f)\}$, which will be then processed through the frozen $\mathcal{F}$.

\paragraph{Training Details}
To validate the effectiveness of our method, we compare the performance of \textsc{PromptFL} with existing framework by 1)training the collaborative model from the scratch and 2)fine-tuning the full model with pretrained weights. We evaluate the performance across four representative dataset used in CLIP for both general objects and fine-grained classification. We report the performance with two representative and influential backbone, Resnet50(38.3M parameters) and Vit\underline{\space}b16(86.6M parameters). For the evaluation metrics, we select three aspects to assess the performance of each method, 1)representative Top-1 accuracy on the test set, 2)F1 score to measure the weighted and unified average of precision and recall, which is more useful especially on unbalanced class distribution, 3)as well as the computational and communication cost reported in Fig.~\ref{fig:cost}. We presuming that higher result on accuracy and F-1 score as well as lower result on computation latency will lead to better a framework, detailed comparison results show the superior if \textsc{PromptFL} in Tab.~\ref{tab:mainresult}. 

All experiments are conducted with Pytorch on GeForce RTX 3090 GPU. Training is performed with SGD with 0.001 learning rate. Tab.~\ref{tab:mainresult} measures the overall performance of \textsc{PromptFL} against existing framework from the perspective of two data distribution settings. For the iid setting, each client shares the same classes, while for the extreme non-iid setting, each client owns the independent and non-overlapping classes. We can see that from Tab.~\ref{tab:mainresult}, \textsc{PromptFL} obtains superior results with similar or better accuracy and F1 value, but with only 0.01\% $\sim$ 0.1\% learnable parameters with the iid setting. Further more, with the non-iid setting \textsc{PromptFL} achieves competitive performance on both accuracy and efficiency against existing framework. Superior outcome on both settings manifest the advantage of our proposed \textsc{PromptFL}. 

\paragraph{Data Distribution Analysis}
After obtaining the decent performance in both extreme iid and non-iid setting, we hope to further testify the stability of \textsc{PromptFL} and figure out the impact of different data distribution on clients to the performance of \textsc{PromptFL}. To observe the intermediate status, we select $p$\% overlapped ratio of classes from 0\% to 10\%, 20\% and 50\%, which means that $p$\% of classes will appear on more than one client and the remaining $1-p$\% classes only shows on single client. Fig. ~\ref{fig:repeat} reports the accuracy and F1 with corresponding distribution. From the result, we observe that given the circumstance that the class on each client is sufficient, the distribution of class has not much impact on the performance of \textsc{PromptFL}, only a tiny improvement when the overlapping of classes reaches 50\%. On the contrary, existing framework shows miserable stability when encountering shifted class distribution other than unified mode by observing the Tab. ~\ref{tab:mainresult}

\paragraph{Impact of number of shots}
Following the few-shot evaluation setting adopted in CLIP, we use 2, 4, 8, 16 shots in training \textsc{PromptFL} and validate the performance with corresponding test sets. Unlike the circumstance in the centralized mode where data only from a single entrance, \textsc{PromptFL} involves several participants. Thus we redeclare that the number of shots for the FL mode implies the overall shot containing the entire participants, which is more practical and accord with the class unbalance scenarios in the real-world. From the result in ~\ref{fig:line}, we observe that as the number of training examples per class increases, the performance of \textsc{PromptFL} enhanced. Furthermore, for the setting of adequate clients with sufficient class number on each client, the accuracy for each setting reveals rather steady. 

\paragraph{Comparison with different clients}
Next, to eliminate the possible impact caused by different clients, we further study the performance of \textsc{PromptFL} with different clients from 16 to 32 to 64, with the iid mode that each client owns random set of classes. Also, to avoid that the collapse of performance due to deficiency of class on each client, especially for the case with large clients number, we also range the number of shots from 2 to 4 to 8 to 16. We observe that for different number of clients, performance will reach similar optimum for as the classes on each client is sufficient. For example, in caltech101, all settings achieve around 89\% with 16 shots.

\paragraph{Computation and Communication Cost Analysis}
We also analyse the efficiency of \textsc{PromptFL} with regard to the computation and communication cost during training. We measure the communication cost by the size of uploaded data per round, and the total round to be transmitted. For the computation cost, we calculate the GPU memory utilization and training GPU time for given steps. Fig. ~\ref{fig:cost} shows the comparison between existing finetuning framework and our proposed \textsc{PromptFL}. We observe that \textsc{PromptFL} can save at most $110$ times communication cost per round compared to existing prevailing method, let alone that \textsc{PromptFL} takes half of rounds to reach convergence, which makes a wider disparity in communication cost between them. As for the computation cost, we report the comparison of GPU time as in the same given steps, where \textsc{PromptFL} remains outperform existing framework around $3$ times. Further more, there is huge advantage that \textsc{PromptFL} consumes far less GPU memory during training, which can alleviate the system burden in practical.

\section{Conclusion}
Overall, there are many unknowns about \textsc{PromptFL} and this paper sets out to investigate its feasibility.
In summary: (1) We demonstrate the system feasibility of \textsc{PromptFL} on modern hardware, in terms of overhead in communication, training, and inference.
(2) We show that \textsc{PromptFL} keeps data on each device private, aiming to learn global prompts updated only by communicating gradients rather than the data itself, and thus not less private than FL.
(3) We implement a proof-of-concept in the framework, spanning a range of popular image classification tasks. We find \textsc{PromptFL} to be competitive with strong FL baselines.

\pagebreak
\newpage

\bibliography{anonymous-submission-latex-2023}
\end{document}